# Slope Stability Analysis with Geometric Semantic Genetic Programming

**Juncai Xu[1,2], Zhenzhong Shen[1], Qingwen Ren[1], Xin Xie[3], and Zhengyu Yang[3]**

*1. College of Water Conservancy and Hydropower Engineering, Hohai University, Nanjing, China*

*2. State Key Laboratory of Simulation and Regulation of Water Cycle in River Basin, China Institute of Water Resources and Hydropower Research, Beijing, China*

*3. Department of Electrical and Computer Engineering, Northeastern University, Boston, MA USA*

E-mail: *xujc@hhu.edu.cn*

**Abstract**:
Genetic programming has been widely used in the engineering field. Compared with the conventional genetic programming and artificial neural network, geometric semantic genetic programming (GSGP) is superior in astringency and computing efficiency. In this paper, GSGP is adopted for the classification and regression analysis of a sample dataset. Furthermore, a model for slope stability analysis is established on the basis of geometric semantics. According to the results of the study based on GSGP, the method can analyze slope stability objectively and is highly precise in predicting slope stability and safety factors. Hence, the predicted results can be used as a reference for slope safety design.

# 1 Introduction

Slope stability is an important issue in slope safety evaluation methods. At present, the widely used evaluation methods include the limit equilibrium method and numerical analysis method [1, 2, 17-26]. Although these methods are supported by the perfect theories of mechanics and scientific systems, these methods are nonetheless based on assumptions. Hence, given the nonlinear relations between various factors that influence slope systems, these methods are ineffective in predicting slope stability. These methods also have other weaknesses, such as a complicated computing process and large computing scale.

To address the aforementioned problems, machine learning algorithms, including neural net algorithm, genetic programming, and adaptive fussy inference, are applied in slope stability prediction. Studies have demonstrated that these artificial intelligent methods can be used in slope stability analysis [3, 4, 27-40]. Despite these methods, some problems remain. For example, when the neural net algorithm is adopted for prediction, slow convergence speed, local convergence, and overfitting usually occur because the relationship between the input and output variables are considered one black box. Although genetic programming can solve the formula for slope stability prediction, the problem of low convergence rate and divergent results still exist, except at a large computing scale [5–7]. A support vector machine (SVM) based on minimum risk probability and a least square SVM (LSSVM) were recently applied in slope stability analysis. According to the analysis results, SVM is relatively effective in predicting safety factors but is deficient in classifying the stability state of slopes. Thus, we use a geometric semantic algorithm to improve computing efficiency and reduce the errors by improving the evolution strategy in genetic programming. As revealed by the application of GSGP in the prediction of the strength of high-performance concrete and the effect of drugs, GSGP can solve regression problems well [8–11]. Thus, to further investigate the slope stability analysis, we introduce GSGP. However, such analyses involve not only the safety factor regression problem but also the classification of slope stability status; thus, the model based on GSGP must also solve the classification problem. Evidently, this research is significant in addressing the slope stability analysis problem and in developing GSGP theory.

# 2 Geometric Semantic Genetic Programming Algorithm

## 2.1 Genetic Programming Regression Algorithm

Genetic programming algorithms are based on a structuring process by using an evolutionary function. This algorithm uses genetic manipulation, including reproduction, crossover, and mutation, to derive the solution of different iterations. The optimal solution is retained as the result of the problem. The solution of genetic programming can be represented by a one-tree structure–function expression, which is often considered the function and terminator sets. However, the conventional genetic programming algorithm does not consider the actual meaning of the function. The semantic genetic programming algorithm was developed from the conventional genetic programming algorithm. For crossover and mutation operations, the GSGP algorithm uses the geometric semantic approach instead of the binary tree of the conventional genetic algorithm.

The specific implementation steps of the GSGP are as follows (Fig.1):

(1) In the initialization process, individuals consist of the function set $F$ and terminator set $T$ to generate the initial population. The function and terminator sets are expressed as follows:

$$F = \{f_1, f_2, \cdots, f_n\}, \quad (1)$$

where $f_i$ denotes the mathematical operation symbols, including $+, -, \times, \div$.

$$T = \{t_1, t_2, \cdots, t_n\} \quad (2)$$

where $t_i$ is the variable comprising the terminator set.

(2) The fitness function is used to evaluate the quality standards for each individual in the population and to calculate the fitness of each individual in the population. This function is also the driving process of the evolution to assess each individual. The degrees of measurement for adaptation methods usually include original fitness, fitness, and standard normalized fitness. When using the error index, the original fitness can be defined as follows:

$$e(i,t) = \sum_{j=1}^{n} |s(i,j) - c(j)| \quad (3)$$

where $s(i,j)$ is the computational result of individual $i$ in instance $j$, $n$ is the number of instances, and $c(j)$ is the actual result of instance $j$.

(3) The genetic operation consists of a parent individual copy, crossover, and mutation operations. By using the parent individual, which is produced by using the geometric semantic method, the crossover operation generates individual $T_C$ and the mutation operation generates individual $T_M$, which are expressed as follows:

$$T_c = (T_1 \cdot T_R) + (1 - T_R) \cdot T_2 \quad (4)$$

where $T_1$ and $T_2$ are two parent individuals, and $T_R$ is a real random number.

$$T_M = T + ms \cdot (T_{R1} - T_{R2}) \quad (5)$$

where $T$ is the parent individual, $T_{R1}$ and $T_{R2}$ are two real random numbers, and $ms$ is the mutation step.

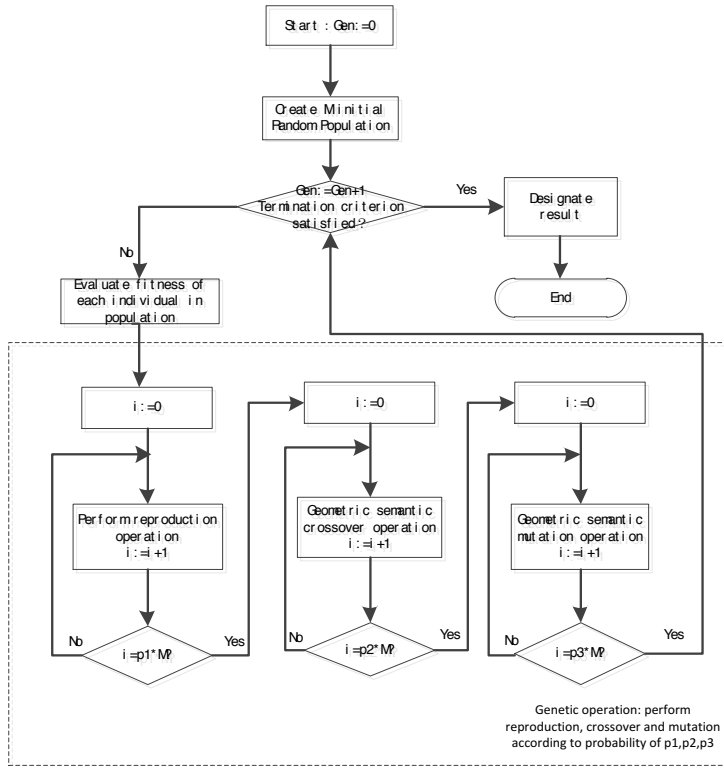

Fig. 1. Operational structure of the GSGP.

### 2.2 Genetic Programming Classification Algorithm

In sample classification by using genetic programming, the operations for achieving prediction results include copying, exchange, and mutation. Unlike in regression problems, all category boundaries have to be determined in classification problems and the fitness function is defined by the classification correctness of [13]. For a multi-category problem, we need to determine multiple boundaries. By contrast, the bi-class classification problem requires the determination of only one cut-off point. When the objection function value is beyond the cut-off point, the result is set to 1; otherwise, the result is set to −1.

In the classification problem under genetic programming, the fitness can be defined as follows [14]:

$$fitness = 1 - \frac{R_{num}}{S_{num}}  \qquad (6)$$

where $R_{num}$ is the number of times that the individual classification is correct and $S_{num}$ is the number of classification samples. The fitness, which is the classification error, is reflected by the sample classification accuracy.

## 3  Slope Stability Analysis Model for Genetic Programming

According to references and conclusions, the effect factors for the slope stability analysis problem includes unit weight (γ), cohesion (*c*), angle of internal friction (*Φ*), slope angle (β), slope height (H), and pore water pressure coefficient ($r_u$) as input variables [15]. Either the status of slope (S) or factor of safety (FS) is included as the output variable. The network structure is composed of six input units and one output unit. Under GSGP, given that the slope stability state can only either be stable or unstable, we can consider the problem as a binary classification problem. We take the result as 1 when the slope is stable; otherwise, we take the result as −1. Fitness is the classification error of the slope stability state. However, under GSGP, the prediction of slope safety factor can be considered a typical nonlinear regression problem. Therefore, in the solution process, the prediction error of slope safety factor can be directly defined as fitness.

A series of evolution parameters for the GSGP has to be determined in the solution process. In the established model, the function set is $\{+,-,\times,\div\}$ and the terminator set has eight variables $\{x_1, x_2, \cdots, x_6\}$ corresponding to $\gamma, c, \Phi, \beta, H, r_u$. Furthermore, we set the size of the population, iterative algebraic algorithm, and variation coefficient according to the computational conditions of the genetic programming algorithm. Given the training samples, the GSGP algorithm can predict the recycled concrete slump of the test samples.

## 4 Engineering Application and Effect Analysis

### 4.1 Application Engineering

By using a slope experimental dataset for testing (Table 1) [6, 16], the established model is applied to predict the slope status and safety factors.

Table 1. Dataset used in the study

| No. | Input variable | | | | | | Actual | | Computational | |
|---|---|---|---|---|---|---|---|---|---|---|
| | $\gamma$ (kN/m) | $c$ (kPa) | $\Phi$ (°) | $\beta$ (°) | $H$ (m) | $r_u$ | S | FS | S | FS |
| 1 | 18.80 | 14.40 | 25.02 | 19.98 | 30.6 | 0 | 1 | 1.876 | -1 | 1.473 |
| 2 | 18.77 | 30.01 | 9.99 | 25.02 | 50 | 0.1 | 1 | 1.400 | 1 | 1.313 |
| 3 | 19.97 | 19.96 | 36 | 45 | 50 | 0.5 | -1 | 0.829 | -1 | 0.963 |
| 4 | 22.38 | 10.05 | 35.01 | 45 | 10 | 0.4 | -1 | 0.901 | -1 | 0.890 |
| 5 | 18.77 | 30.01 | 19.98 | 30 | 50 | 0.1 | 1 | 1.460 | 1 | 1.359 |
| 6 | 28.40 | 39.16 | 37.98 | 34.98 | 100 | 0 | 1 | 1.989 | 1 | 2.000 |
| 7 | 19.97 | 10.05 | 28.98 | 34.03 | 6 | 0.3 | 1 | 1.340 | 1 | 1.257 |
| 8 | 13.97 | 12.00 | 26.01 | 30 | 88 | 0 | -1 | 1.021 | -1 | 0.848 |
| 9 | 18.77 | 25.06 | 19.98 | 30 | 50 | 0.2 | -1 | 1.210 | -1 | 1.213 |
| 10 | 18.83 | 10.35 | 21.29 | 34.03 | 37 | 0.3 | -1 | 1.289 | -1 | 1.227 |
| 11 | 28.40 | 29.41 | 35.01 | 34.98 | 100 | 0 | 1 | 1.781 | 1 | 1.673 |
| 12 | 18.77 | 25.06 | 9.99 | 25.02 | 50 | 0.2 | -1 | 1.180 | -1 | 1.173 |
| 13 | 16.47 | 11.55 | 0 | 30 | 3.6 | 0 | -1 | 1.000 | -1 | 0.982 |
| 14 | 20.56 | 16.21 | 26.51 | 30 | 40 | 0 | -1 | 1.250 | -1 | 1.199 |
| 15 | 18.66 | 26.41 | 14.99 | 34.98 | 8.2 | 0 | -1 | 1.111 | -1 | 1.154 |
| 16 | 13.97 | 12.00 | 26.01 | 30 | 88 | 0.5 | -1 | 0.626 | -1 | 0.848 |
| 17 | 25.96 | 150.1 | 45 | 49.98 | 200 | 0 | 1 | 1.199 | 1 | 1.271 |
| 18 | 18.46 | 25.06 | 0 | 30 | 6 | 0 | -1 | 1.090 | -1 | 1.059 |
| 19 | 19.97 | 40.06 | 30.02 | 30 | 15 | 0.3 | 1 | 1.841 | 1 | 1.956 |
| 20 | 20.39 | 24.91 | 13.01 | 22 | 10.6 | 0.4 | 1 | 1.400 | 1 | 1.439 |
| 21 | 19.60 | 12.00 | 19.98 | 22 | 12.2 | 0.4 | -1 | 1.349 | -1 | 1.341 |
| 22 | 20.96 | 19.96 | 40.01 | 40.02 | 12 | 0 | 1 | 1.841 | 1 | 1.786 |
| 23 | 17.98 | 24.01 | 30.15 | 45 | 20 | 0.1 | -1 | 1.120 | -1 | 1.205 |
| 24 | 20.96 | 45.02 | 25.02 | 49.03 | 12 | 0.3 | 1 | 1.529 | 1 | 1.502 |
| 25 | 22.38 | 99.93 | 45 | 45 | 15 | 0.3 | 1 | 1.799 | 1 | 1.838 |
| 26 | 18.77 | 19.96 | 19.98 | 30 | 50 | 0.3 | -1 | 1.000 | -1 | 1.072 |
| 27 | 21.78 | 8.55 | 32 | 27.98 | 12.8 | 0.5 | -1 | 1.030 | -1 | 1.151 |
| 28 | 21.47 | 6.90 | 30.02 | 31.01 | 76.8 | 0.4 | -1 | 1.009 | -1 | 1.007 |
| 29 | 21.98 | 19.96 | 22.01 | 19.98 | 180 | 0.1 | -1 | 0.991 | -1 | 1.006 |
| 30 | 18.80 | 57.47 | 19.98 | 19.98 | 30.6 | 0 | 1 | 2.044 | 1 | 1.930 |
| 31 | 21.36 | 10.05 | 30.33 | 30 | 20 | 0 | 1 | 1.700 | 1 | 1.572 |
| 32 | 18.80 | 14.40 | 25.02 | 19.98 | 30.6 | 0.5 | -1 | 1.111 | -1 | 1.473 |
| 33 | 15.99 | 70.07 | 19.98 | 40.02 | 115 | 0 | -1 | 1.111 | -1 | 1.130 |
| 34 | 21.98 | 19.96 | 36 | 45 | 50 | 0 | -1 | 1.021 | -1 | 1.018 |
| 35 | 19.08 | 10.05 | 9.99 | 25.02 | 50 | 0.4 | -1 | 0.649 | -1 | 0.699 |
| 36 | 19.08 | 10.05 | 19.98 | 30 | 50 | 0.4 | -1 | 0.649 | -1 | 0.754 |
| 37 | 17.98 | 45.02 | 25.02 | 25.02 | 14 | 0.3 | 1 | 2.091 | 1 | 2.009 |
| 38 | 24.96 | 120.0 | 45 | 53 | 120 | 0 | 1 | 1.301 | 1 | 1.273 |
| 39 | 20.39 | 33.46 | 10.98 | 16.01 | 45.8 | 0.2 | -1 | 1.280 | -1 | 1.289 |
| 40 | 17.98 | 4.95 | 30.02 | 19.98 | 8 | 0.3 | 1 | 2.049 | 1 | 1.931 |
| 41 | 18.97 | 30.01 | 35.01 | 34.98 | 11 | 0.2 | 1 | 2.000 | 1 | 1.726 |
| 42 | 21.98 | 19.96 | 22.01 | 19.98 | 180 | 0 | -1 | 1.120 | -1 | 1.006 |
| 43 | 20.96 | 30.01 | 35.01 | 40.02 | 12 | 0.4 | 1 | 1.490 | 1 | 1.492 |
| 44 | 20.96 | 34.96 | 27.99 | 40.02 | 12 | 0.5 | 1 | 1.430 | 1 | 1.487 |
| 45 | 18.46 | 12.00 | 0 | 30 | 6 | 0 | -1 | 0.781 | -1 | 0.996 |
| 46 | 19.97 | 40.06 | 40.01 | 40.02 | 10 | 0.2 | 1 | 2.310 | 1 | 1.935 |

| | | | | | | | | | |
|---|---|---|---|---|---|---|---|---|---|
| 47 | 19.97 | 19.96 | 36 | 45 | 50 | 0.3 | -1 | 0.961 | -1 | 0.963 |
| 48 | 18.77 | 19.96 | 9.99 | 25.02 | 50 | 0.3 | -1 | 0.970 | -1 | 1.011 |
| 49 | 18.83 | 24.76 | 21.29 | 29.2 | 37 | 0.5 | -1 | 1.070 | -1 | 1.200 |
| 50 | 19.03 | 11.70 | 27.99 | 34.98 | 21 | 0.1 | -1 | 1.090 | -1 | 1.199 |
| 51 | 22.38 | 10.05 | 35.01 | 30 | 10 | 0 | 1 | 2.000 | -1 | 1.564 |
| 52 | 18.80 | 15.31 | 30.02 | 25.02 | 10.6 | 0.4 | 1 | 1.631 | 1 | 1.747 |

In model testing, from the dataset of 52 samples shown in Table 1, the first 40 samples are considered the training set and the remaining samples are considered the test set. The size of the initial population is set to 500, and the maximum iteration epoch is set to 50. The genetic mutation step is set to 0.1. By using the GSGP algorithm, we can obtain the slope status and safety factors of the training and testing datasets. The results correspond to the last two columns in Table 1. As shown in the table, the slope status and safety factors of the datasets can be determined when the six parameters ($\gamma$, $c$, $\Phi$, $\beta$, $H$, $r_u$) are given.

### 4.2 Effect Analysis

To evaluate the performance of the GSGP algorithm, the results for predicting the slope status and safety factors need to be analyzed. Slope status prediction is a classification problem. However, safety factor prediction is a regression problem. The resulting classification accuracy and correlation coefficients are used to evaluate the performance of the algorithm.

The formula for the classification prediction accuracy of the GSGP is defined as follows:

$$Accuracy(\%) = \left( \frac{Number\ of\ data\ predicted\ accurately\ by\ GSGP}{Total\ data} \right) \times 100 \quad (7)$$

By using the classification prediction results and actual values in Table 1, the accuracies for the training and testing datasets can be obtained by using Eq. (7). The classification accuracies for the training and testing datasets were 97.5% and 91.7%, respectively. On the basis of the existing references, the accuracy of the GSGP algorithm is superior to that of other algorithms such as the ANN and SVM.

With regard to the safety factor, the index correlation coefficient between the true and computational values is defined as follows [10]:

$$R = \frac{n \sum y \cdot y' - (\sum y)(\sum y')}{\sqrt{\sum y^2 - (\sum y)^2} \sqrt{\sum y'^2 - (\sum y')^2}} \quad (8)$$

where $y$ is the true value, $y'$ is the computational value, and $n$ is the sample size.

In the same way, the correlation coefficient can be obtained in Eq. (8) by using the prediction result for safety factor and the actual values in Table 1. Given the FS of the training and testing datasets, the calculation results are shown in Figures 2–3.

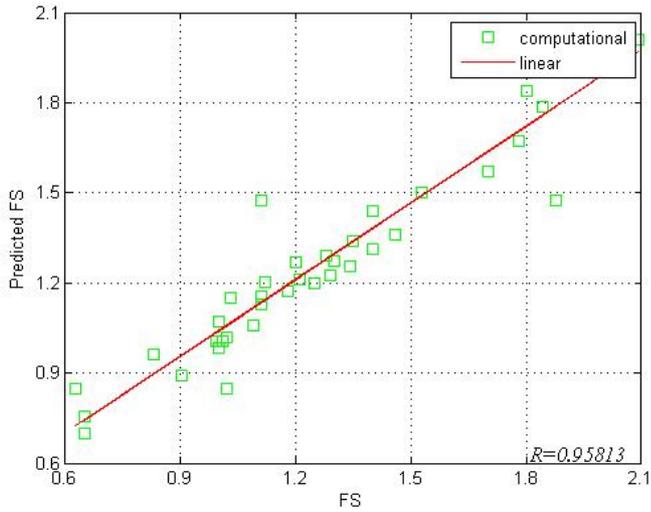

Fig 2. Performance for training dataset using GSGP

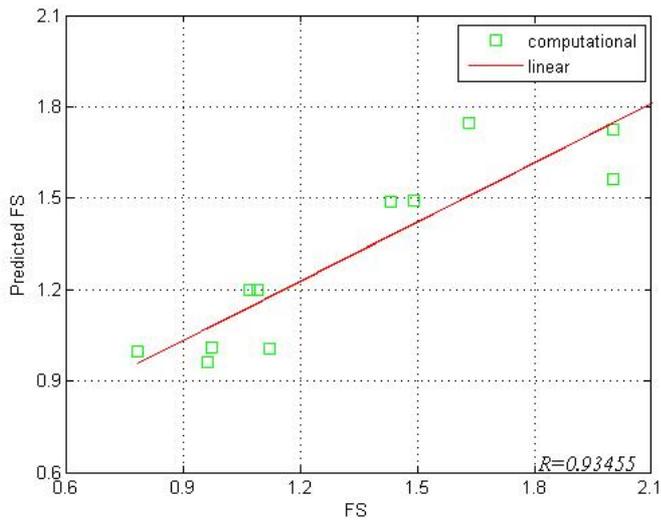

Fig 3. Performance for the testing dataset by using GSGP

In Figures 2–3, the results show that the correlation coefficient of the predicted and actual values of FS for the training dataset was 95.8%. The correlation coefficient of the predicted and actual values of FS for the testing dataset was 93.4%. The precision for the training dataset is slightly higher than that for the testing dataset. Nonetheless, high correlation exists for both datasets.

To further study the accuracy of GSGP, we compared its RMSE with two others algorithm, support vector machine (SVM) and standard genetic programing (STGP). Based on Table 1 data sets, is the three algorithms were run 50 times for the predication of recycled concrete slump. Fig. 4 is the statistical results of RMSE about the three algorithms.

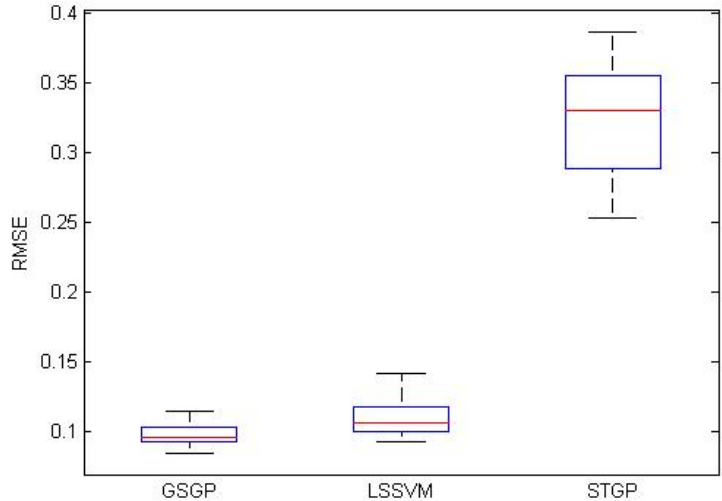

Fig. 4. Box-whisker of three algorithm errors in the test dataset

From Fig. 4, the errors rang, the interquartile ranges (IQR), using GSGP algorithm is narrowest in the three methods, however, the error rang using STGP algorithm is widest. In the Wilcoxon rank-sum analysis, p of the three algorithms, the p from GSGP algorithm is still lowest in three algorithms. Thus, the solutions using GSGP algorithm is significantly better than the other two algorithms.

## 5  Conclusion

Given the variety of complex factors in slope stability analysis, the use of a mechanical method therein is excessively difficult. By using such methods, the prediction for slope status and safety factors often become poor. In this paper, GSGP was introduced into slope stable analysis. The established slope stability analysis method based on the GSGP algorithm successfully predicted slope status and safety factors. Furthermore, upon application of the model in slope stability analysis, the following conclusions were made:

(1) By adapting the GSGP algorithm to different adjustments, the classification or regression analysis of datasets results in different functions;

(2) When the unit weight ($\gamma$), cohesion (c), angle of internal friction ($\Phi$), slope angle ($\beta$), slope height (H), and pore water pressure coefficient ($r_u$) are considered input variables, GSGP can predict the slope status and safety factors;

(3) Within the allowable GSGP prediction error, the method can be applied in slope stability analysis. The results can be used as a reference in the design process for slopes.

**Acknowledgement**

This research was funded by the Open Research Fund of State Key Laboratory of Simulation and Regulation of Water Cycle in River Basin (Grant No. IWHRSKL-201518) and A Project Funded by the Priority Academic Program Development of Jiangsu Higher Education Institutions (Grant No.3014-SYS1401).